\documentclass{article}
\usepackage{spconf,amsmath,graphicx}
\usepackage{amsfonts}
\usepackage{adjustbox}
\usepackage[parfill]{parskip}
\usepackage{mathtools}
\usepackage{xcolor}

\usepackage{graphicx}
\usepackage{subcaption}

\title{Progressive Continual Learning for Spoken Keyword Spotting}
\name{Yizheng Huang$^{\star}$\qquad Nana Hou$^{\dagger}$ \qquad Nancy F. Chen$^{\star}$}
\address{$^{\star}$Institute for Infocomm Research, A*STAR, Singapore \\ $^{\dagger}$Nanyang Technological University, Singapore}
\begin{document}
% \ninept
%
\maketitle
\begin{abstract}
Catastrophic forgetting is a thorny challenge when updating keyword spotting (KWS) models after deployment. To tackle such challenges, we propose a progressive continual learning strategy for small-footprint spoken keyword spotting (PCL-KWS). Specifically, the proposed PCL-KWS framework introduces a network instantiator to generate the task-specific sub-networks for remembering  previously learned keywords. As a result, the PCL-KWS approach incrementally learns new keywords without forgetting prior knowledge. Besides, the proposed keyword-aware network scaling mechanism of PCL-KWS constrains the growth of model parameters while achieving high performance. Experimental results show that after learning five new tasks sequentially, our proposed PCL-KWS approach archives the new state-of-the-art performance of 92.8\% average accuracy for all the tasks on \textit{Google Speech Command} dataset compared with other baselines.
\end{abstract}


\section{Introduction}
\label{sec:intro}

%%%%%%%%%%%%%%%%%%%%%%%%%%%%%%%%%%%%%%%%%%%%%%%%%%%%%%%%%%%%%%%%%%
%%% Introduction
%%%%%%%%%%%%%%%%%%%%%%%%%%%%%%%%%%%%%%%%%%%%%%%%%%%%%%%%%%%%%%%%%%
Spoken keyword spotting (KWS) aims to identify specific keywords in the audio input. It serves as a primary module in many real-word applications, such as Apple Siri and Google Home. Current approaches to keyword spotting are fueled by deep learning models \cite{chen2014small, chen2016exemplar, sainath2015convolutional, chen2014strategies}, which are usually trained with limited keywords in compact model for lower computation and smaller footprint \cite{sainath2015convolutional, chen2015low}. Therefore, the performance of the KWS model trained by the source-domain data may degrade significantly when confronted with unseen keywords of the target-domain at run-time. 

To alleviate such problem, prior work \cite{mazumder2021few, awasthi2021teaching, lin2020training} utilizes few-shot fine-tuning \cite{sun2019meta} to adapt KWS models with training data from the target-domain for new scenarios. The limitations of such an approach are two-fold. First, performance on data from the source domain after adaptation could be poor, which is also known as \textit{catastrophic forgetting} problem \cite{mccloskey1989catastrophic}. Second, the fine-tuning process usually requires a large pre-trained speech embedding model to guarantee the performance, requiring higher memory cost and is undesirable for small-footprint KWS scenarios.

Continual learning (CL) \cite{parisi2019continual, huang2021modelci, zhang2021serverless} addresses the catastrophic forgetting issue when adapting pre-trained model with target-domain data. Some efforts in continual learning have been applied to speech recognition task \cite{chang2021towards, sadhu2020continual}. 
We can categorize existing CL methods based on replay buffer (the requirement of the extra memory to store data/parameters) into two categories: regularization-based methods and replay-based methods. Regularization-based methods protects the parameters learned from previous tasks by loss regularization (e.g., EWC \cite{kirkpatrick2017overcoming}, SI \cite{zenke2017continual}). Replay-based methods require a buffer to store 1) historical data as the replay exemplars (e.g., Rehearsal \cite{hsu2018re, rebuffi2017icarl}, GEM \cite{lopez2017gradient}), or 2) model weights to isolate the parameters learned by different tasks \cite{xu2018reinforced, rosenfeld2018incremental, hou2020multi}.

In this paper, we first investigate and analyze the prior four continual learning strategies (EWC, SI,  NR, GEM) for the keyword spotting task. Then, we propose a progressive continual learning strategy for small-footprint keyword spotting (PCL-KWS) to alleviate the catastrophic forgetting problem when adapting the model trained by source-domain data with the target-domain data. Specifically, the proposed PCL-KWS includes several task-specific sub-networks to memorize the knowledge of the previous keywords. Then, a keyword-aware network scaling mechanism is introduced to reduce the network parameters. As a result, the proposed PCL-KWS could preserve the high accuracy of the old tasks when learning the new tasks without increasing model parameters significantly. The experimental results show that PCL-KWS achieves state-of-the-art performance with only a few extra parameters and no buffers.

%%%%%%%%%%%%%%%%%%%%%%%%%%%%%%%%%%%%%%%%%%%%%%%%%%%%%%%%%%%%%%%%%%
%%% CL Methods
% 1. General continual learning workflow
% 2. Regularization-based methods
% 3. Replay methods
% 4. *Progressive TC-ResNet (our method, name can be changed)
%%%%%%%%%%%%%%%%%%%%%%%%%%%%%%%%%%%%%%%%%%%%%%%%%%%%%%%%%%%%%%%%%%
\section{PCL-KWS Architecture}
\label{sec:method}

This section first describes four continual learning strategies, including regularization-based methods and reply-based methods. The proposed progressive continual learning for keyword spotting (PCL-KWS) are then introduced.

\subsection{Continual Learning Workflow}
The workflow of conducting continual learning for keyword spotting is to receive $T$ landed keyword learning tasks sequentially and optimize performances on all the $T$ tasks without catastrophic forgetting. For each task $\tau_t$ with $t \le T$, we have a training dataset including pairs $(x_t, y_t)$, where $x_t$ are the audio utterances following the distribution $D^T$ and $y_t$ are the keyword labels. We aim to minimize the total loss function $\mathcal{L}_{tot}$, which sums the sub-losses $\mathcal{L}_{kws}$ of all $T$ tasks formulated as:
\begin{equation}
\label{eq:cl}
\mathcal{L}_{tot} = \sum_{t=0}^{T}\mathbb{E}_{(x_t, y_t)\sim D^T}\left [ \mathcal{L}_{kws}(F_t(x_t; \theta^t), y_t) \right ]
\end{equation}
where $ \mathcal{L}_{kws} $ denotes the cross-entropy loss and $F_t(x_t; \theta^t)$ is the KWS model with parameters $\theta^t$. During training, we seek the optimal parameters $\theta^{t'}$ for all the $T$ tasks. It is a challenging task since the parameters learned from the previous task $\tau_t$ are easily overwritten after learning the new task $\tau_{t+1}$, incurring catastrophic forgetting.

To robustify performance, we introduce regularization-based methods and replay-based methods for the KWS task.

\subsubsection{Regularization-based Methods}
The regularization-based methods protect important parameters trained on previous tasks from being overwritten by adding regular terms to the loss function $\mathcal{L}_{kws}$. We introduce two commonly-used regularization terms: Elastic Weight Consolidation (EWC) \cite{kirkpatrick2017overcoming} and Synaptic Intelligence (SI) \cite{zenke2017continual}, which are explored as baselines in this work.

\textbf{EWC} first calculates the parameter importance $\Omega_i^t$ for $i^{th}$ parameter of $\theta^t$ in each task $\tau_t$. $\Omega_i^t$ acts as the importance regularization to constrain more important parameters as close to those of previous $t-1$ tasks, while updating parameters of less importance with the $t^{th}$ new task. The optimized loss function is formulated as:
\begin{equation}
\label{eq:ewc}
\mathcal{L}_{kws}' = \mathcal{L}_{kws}(F_t(x_t; \theta^t), y_t) + \frac{\lambda}{2} \sum_{i}\Omega_i^t(\theta_i^t - \theta_i^{t - 1})^2, 
\end{equation}
where the importance coefficient $\Omega_i^t$ is determined by the fisher information matrix (FIM):
\begin{equation}
\label{eq:ewc-2}
\Omega_i^t = \mathbb{E}_{(x_t, y_t)\sim D^T}\left [ \left ( \frac{\delta\mathcal{L}_{kws}}{\delta\theta_i^t} \right )^2 \right ].
\end{equation}

\textbf{SI} is derived from EWC, which assigns the parameter importance $\Omega_{i}^t$ in a different manner. It utilizes the gradient $\omega^t_i$, calculated from the KWS loss $\mathcal{L}_{kws}$, to estimate the $i^{th}$ parameter importance at each training step, 
\begin{equation}
\label{eq:si}
\Omega_i^{t+1} = \Omega_{i}^{t} + \frac{\omega^t_i}{\left ( \theta_i^t - \theta_i^{t - 1} \right )^2 + \varepsilon }
\end{equation}
where the damping parameter $\varepsilon$ is used to avoid division by zero, making the training process more stable. 

\subsubsection{Replay-based Methods}
The replay-based methods utilize an in-memory buffer to store the historical training samples to maintain the accuracy of the previous $t-1$ tasks. We adopt Naive Rehearsal (NR) \cite{hsu2018re} and Gradient Episodic Memory (GEM) \cite{lopez2017gradient} to demonstrate replay-based methods.

\textbf{NR} stores the randomly selected training samples from previous $t-1$ learned task $ \left \{ \tau_0, \tau_1 ... \tau_{t-1} \right \}$ into replay buffer, and builds the training data $D_{t}'$ of the task $\tau_t$ formulated as:
\vspace{-2pt}
\begin{equation}
\label{eq:dr}
D_{t}' = \xi({D_1, D_2,..., D_{t - 1})} \cup D_t, \quad 0 < \xi \le 1
\end{equation}
\vspace{-2pt}
where the $\xi\%$ denotes the percentage of utilized historical data, and $D_t$ is the incoming dataset for the new task. 

\textbf{GEM} calculates the gradients $g_k$ on the buffered data of task $\tau_k$ with $k \le t$. If the current gradient $g_t$ of the task $\tau_t$ degrades the performances on any previous tasks, it projects $g_t$ to the gradient $\tilde{g}$, which has the minimum L2 distance to $g_t$. The learning process is formulated as:
\vspace{-2pt}
\begin{alignat}{2}
\label{eq:gem}
    & \underset{\tilde{g}}{\text{minimize}} \ \frac{1}{2}\left \| g_t - \tilde{g} \right \|_2^2 \\
    & \text{subject to} \left \langle \tilde{g}, g_k \right \rangle \geq 0, \text{for all} \ k < t,
\end{alignat}
\vspace{-2pt}
where $\left \langle \cdot ,\cdot \right \rangle$ means the inner product. Positive inner products indicate that gradients $g_k$ and $\tilde{g}$ are in the same direction. The updated weights by such gradients could avoid forgetting the knowledge of previous $t-1$ tasks.

\begin{table*}[t]
\centering
\begin{adjustbox}{max width=0.9\textwidth}
\begin{tabular}{l|cccc|ccc}
\hline
Method             & ACC                     & LA    & BWT   & $\text{ACC}^+$ & TT (sec)            & Extra Param  & Buffer Size  \\ \hline
(a) Fine-tune (lower-bound)        & $0.391_{\ \pm \ 0.265}$ & $0.967_{\ \pm \ 0.015}$ & $-0.264_{\ \pm \ 0.103}$ & -          & 109.24 & -   & -       \\ \hline
(b) EWC                & $0.386_{\ \pm \ 0.268}$                   & $0.774_{\ \pm \ 0.128}$ & $-0.218_{\ \pm \ 0.104}$ & -0.5\%         & 187.27                 &  67.69K   &   -     \\
(c) SI                 & $0.518_{\ \pm \ 0.281}$ & $0.875_{\ \pm \ 0.032}$ & $-0.158_{\ \pm \ 0.025}$ & +12.1\%          & 129.10 &  67.69K   &   -     \\ \hline
(d) GEM-128            & $0.622_{\ \pm \ 0.154}$ & $0.939_{\ \pm \ 0.023}$ & $-0.128_{\ \pm \ 0.033}$ & +23.1\%          & 291.75 & -    & 4.10M   \\
(e) GEM-512            & $0.704_{\ \pm \ 0.117}$ & $0.934_{\ \pm \ 0.019}$ & $-0.115_{\ \pm \ 0.054}$ & +31.3\%          & 318.98 & -    & 15.98M  \\
(f) GEM-1024          & $0.705_{\ \pm \ 0.109}$ & $0.932_{\ \pm \ 0.020}$ & $-0.101_{\ \pm \ 0.036}$ & +31.4\%          & 499.86 & -    & 32.43M  \\
(g) NR ($\xi=0.5$)    & $0.813_{\ \pm \ 0.007}$ & $0.930_{\ \pm \ 0.004}$ & $+0.001_{\ \pm \ 0.012}$ & +42.2\%  &  583.84        & -    & 0.56G \\ 
(h) NR ($\xi=0.75$)    & $0.841_{\ \pm \ 0.011}$ & $0.941_{\ \pm \ 0.010}$ & $\textbf{+0.002}_{\ \pm \ 0.001}$ & +45.0\%  &  721.49         & -    & 0.97G \\ \hline
(i) PCL-KWS (fix)       & $\textbf{0.928}_{\ \pm \ 0.019}$                   & $\textbf{0.949}_{\ \pm \ 0.017}$ & $-0.012_{\ \pm \ 0.011}$          & \textbf{+53.7\%}                &  106.23   & 172.45K  & -     \\
(j) PCL-KWS             & $0.884_{\ \pm \ 0.037}$                   & $0.944_{\ \pm \ 0.022}$ & $-0.020_{\ \pm \ 0.022}$ & +49.3\%          & \textbf{98.41}                  & \textbf{17.46K}    & -       \\ \hline     
(k) Stand-alone (upper-bound) & $0.943_{\ \pm \ 0.003}$ & $0.963_{\ \pm \ 0.247}$ & $+0.002_{\ \pm \ 0.001}$ & +55.2\%          & 123.08                  & 617.87K    & -       \\ \hline
\end{tabular}
\end{adjustbox}
    \caption{\textit{The overall performance of various continual learning strategies on the TC-ResNet-8 model. PCL-KWS (fix) denotes the proposed PCL-KWS without the keyword-aware network scaling mechanism.}}
\label{table:perf}
\end{table*}

\subsection{Proposed Progressive Continual Learning}
%% Nancy: Minimize the usage of subjective adjectives. Anyone can claim their approach is novel. Is facts to demonstrate the novelty instead.  
%%%%%%%%%%%%%%%%%%%%%%%%%%%%%%%%%%%%%%%%%%%%%%%%%%%%%%%%%%%%%%%%
%%%%%%%%%%%%%%%%%% TC-PNN Structure Figure %%%%%%%%%%%%%%%%%%%%%
%%%%%%%%%%%%%%%%%%%%%%%%%%%%%%%%%%%%%%%%%%%%%%%%%%%%%%%%%%%%%%%%
\begin{figure}[t]
  \centering
  \includegraphics[width=0.9\linewidth]{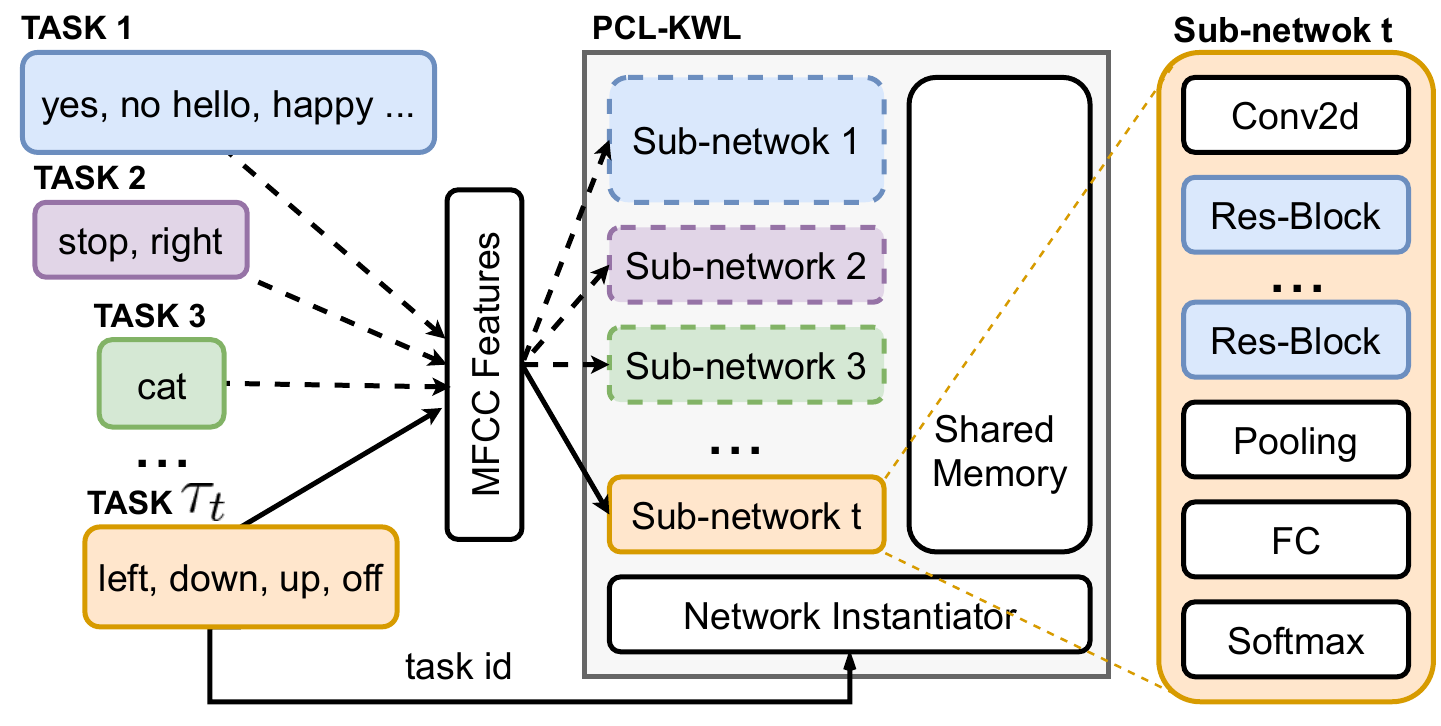}
    \caption{\textit{Block diagram of the proposed PCL-KWS.}}
  \label{fig:net-str}
\end{figure}

Inspired by progressive neural networks \cite{rusu2016progressive} and meta lifelong learning \cite{wang2021mell}, we propose a progressive continual learning strategy for the small-footprint keyword spotting task (PCL-KWS), as shown in Figure \ref{fig:net-str}. 

The proposed PCL-KWS approach consists of two components: a network instantiator and a shared memory connected to a set of classification sub-networks. Specifically, we extract the mel-frequency cepstral coefficients (MFCCs) for several audio utterances in a new learning task $\tau_{t}$ as the inputs. Meanwhile, the proposed PCL-KWS freezes the $t-1$ sub-nets generated for previous $t-1$ tasks. Next, the network instanatiator generates a new $t^{th}$ classification sub-network, which is trained with the MFCC features and metadata of the task (e.g., task id) to identify unseen keywords in the new task $\tau_{t}$. The newly generated $t^{th}$ sub-network shares the same memory along with previous sub-networks, which includes all learned knowledge. As a result, the proposed PCL-KWS can perform knowledge transfer among tasks with faster convergence speed. During inference at run-time, the proposed PCL-KWS selects the corresponding $t^{th}$ sub-network according to the given task $\tau_{t}$ for evaluation.

%%%%%%%%%%%%%%%%%%%%%%%%%%%%%%%%%%%%%%%%%%%%%%%%%%%%%%%%%%%%%%%%%%%%%%%%%%%%%%%
%%%%%%%%%%%%%%%%%% Continual Learning Performance Table %%%%%%%%%%%%%%%%%%%%%%%
%%%%%%%%%%%%%%%%%%%%%%%%%%%%%%%%%%%%%%%%%%%%%%%%%%%%%%%%%%%%%%%%%%%%%%%%%%%%%%%
\textbf{A network instantiator} is designed to generate $t^{th}$ network for each new task $\tau_t$. When a new keyword incremental task $\tau_t$ comes, the instantiator adds a single-head classification sub-network according to the number of the new keywords. Different from the pre-trained model including \{16, 24, 32, 48\} channels, each sub-network in PCL-KWS has \{16, 48\} channels. To constrain the growth of the model parameters, we propose a keyword-aware network scaling mechanism, which multiplies the channels \{16, 48\} with a dynamic width multiplier $\alpha$ \cite{howard2017mobilenets}. The $\alpha_\tau$ for task $\tau_t$ is formulated as,
\begin{equation}
\label{eq:multiplier}
\alpha_{\tau} = \mu\frac{C_t}{C_0}, (\mu > 0),
\end{equation}
\noindent where the $C_t$ and $C_0$ denote the numbers of keywords in the task $\tau_t$ and task $\tau_0$, respectively. The task $\tau_0$ contains the data utilized to pre-train the model. In practice, the incremental task usually has less categories than the pre-trained model, leading to an $\alpha \le 1$ that makes the network smaller.

\textbf{A shared memory} is designed for storing all the learned features of previous $t-1$ task, which can be accessed when learning the new task $\tau_t$.

% It nimbly generates sub-networks (column) dedicated to each incoming keyword learning tasks, and thus alleviates the catastrophic forgetting caused by knowledge overwritten. By designing a column instantiator, we constraint the model size with only a few parameters added, but could remember all the previous knowledge.
%%%%%%%%%%%%%%%%%%%%%%%%%%%%%%%%%%%%%%%%%%%%%%%%%%%%%%%%%%%%%%%%%%
%%% Experiments
%%%%%%%%%%%%%%%%%%%%%%%%%%%%%%%%%%%%%%%%%%%%%%%%%%%%%%%%%%%%%%%%%%
\section{Experiments And Results}
\label{sec:experiment}
\subsection{Dataset} 
We conduct experiments on the \textit{Google Speech Command} dataset (GSC) \cite{warden2018speech}, which includes 64,727 one-second utterance clips with 30 English keywords categories. Following the guideline in \cite{warden2018speech}, we first process all the data with a sample rate of 16kHz. We then split the dataset into two subsets, 80\% for training and 20\% for testing, respectively. 

%%%%%%%% Experimental Setup %%%%%%%%%%%
\subsection{Experimental Setup}

\textbf{Testbed model.} We employ the TC-ResNet-8 \cite{choi2019temporal} as our testbed to explore the various continual learning strategies. It includes three residual blocks with four temporal convolution \cite{hao2020time} layers and each layer contains \{16,24,32,48\} channels. To make a fair comparison, we initialize PCL-KWS using the same settings. 

\textbf{Reference baselines.} We built six baselines for comparisons. The \textit{fine-tune} training strategy adapts the TC-ResNet-8 model for each new task without any continual learning strategies, as the lower-bound baseline. The \textit{stand-alone} strategy trains the TC-ResNet-8 model for each task $\tau_t$ of the dataset separately and could obtain $T$ separate models for evaluation. The four prior continual learning strategies (i.e., EWC, SI, GEM, NR) are introduced in Section \ref{sec:method}. Specifically, at the training stage of naive rehearsal (NR), we replay the training samples from all previous tasks with the percentage $\xi = \{0.5,0.75\}$. At the training stage of gradient episodic memory (GEM), we set the buffer size to $\{128, 512, 1024\}$ training samples.

\textbf{CL setting.} We pre-train the TC-ResNet-8 model on the GSC dataset with 15 randomly selected keywords. To evaluate the baselines along with the proposed PCL-KWS model, we simulate the incremental learning process with 5 different tasks, where each includes 3 new unique keywords. For all methods, we evaluate them on an AWS \textit{t2.medium} instance with only 2 Intel Xeon virtual CPU cores and 4G RAM.

%%%%%%%%%%%%%%%%%%%%%%%%%%%%%%%%%%%%%%%%%%%%%%%%%%%%%%%%%%%%%%%%
%%%%%%%%%%%%%%%%%% Learning Curve Figure %%%%%%%%%%%%%%%%%%%%%%%
%%%%%%%%%%%%%%%%%%%%%%%%%%%%%%%%%%%%%%%%%%%%%%%%%%%%%%%%%%%%%%%%

\subsection{Evaluation Metrics} 
We report performance using the accuracy and efficiency metrics. The accuracy metrics include \textit{average accuracy (ACC)}, \textit{learning accuracy (LA)} \cite{riemer2018learning}, and \textit{backward transfer (BWT)} \cite{lopez2017gradient}. Specifically, 1) \textit{ACC} is the accuracy averaged on all learned tasks, and \textit{$\text{ACC}^+$} is the ACC improvement compared to the fine-tuning baseline. 2) \textit{LA} denotes the accuracy evaluated on the current task. 3) \textit{BWT} evaluates accuracy changes on all previous tasks after learning a new task, indicating the forgetting degree.

The efficiency metrics include \textit{task training time (TT)}, \textit{extra parameters (Extra Param)}, and \textit{buffer size}. 1) \textit{TT} is the per-epoch average training time of all the tasks. 2) \textit{Extra Param} measures the additional parameters brought by the continual learning (CL) strategy. There are no extra parameters for reply-based methods (i.e., NR, GEM). (3) \textit{Buffer size} indicates the extra in-memory cache occupied by the CL strategy for storing replay data or model weights.

\begin{figure}[t]
  \centering
  \includegraphics[width=1.0\linewidth]{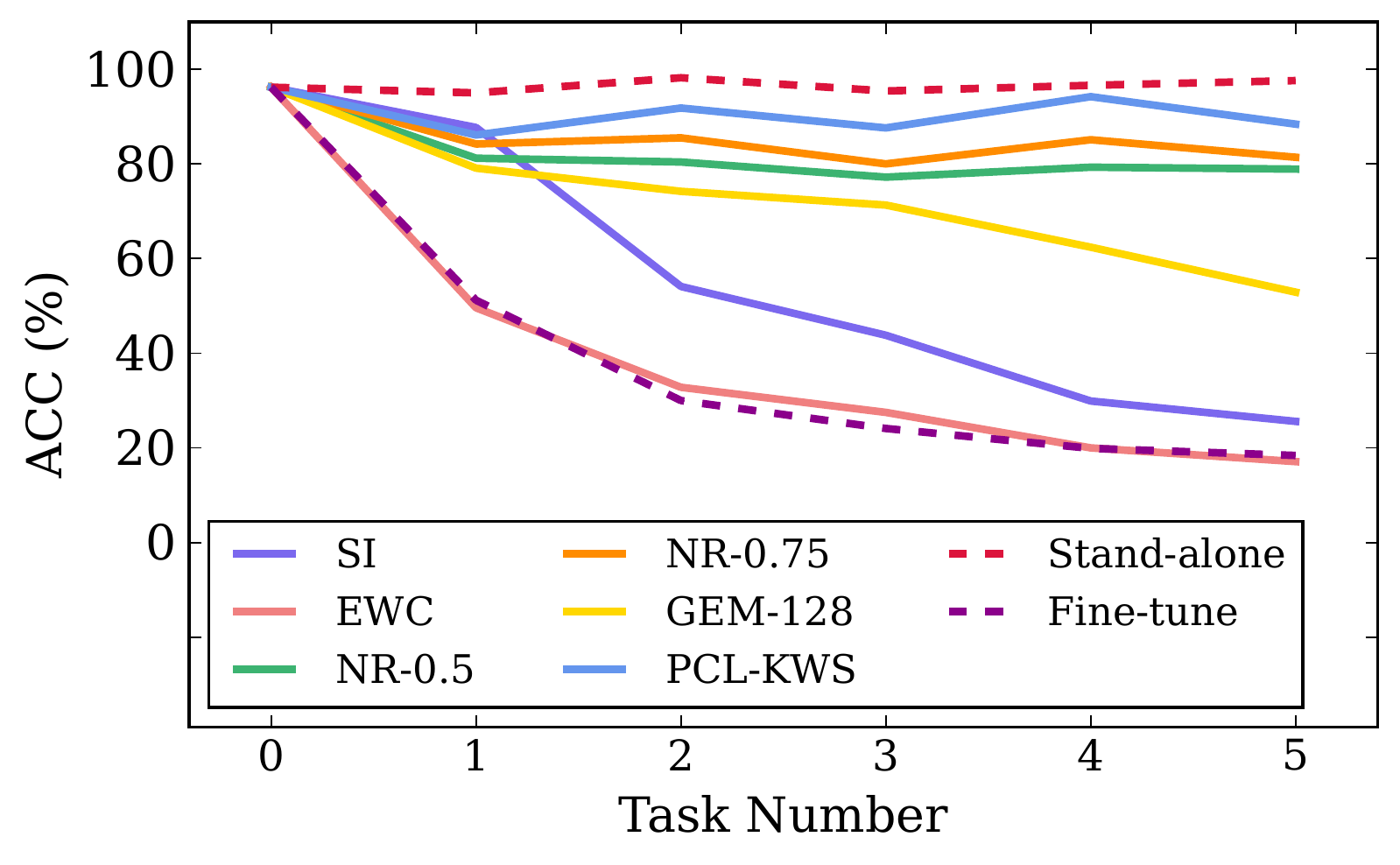}
  \caption{\textit{The ACC (\%) with the number of learned tasks.}}
  \label{fig:learning-curve}
\end{figure}

%%%%%%%% Results and Analysis %%%%%%%%%%%
\subsection{Effect of the regularization-based methods}
Table \ref{table:perf} shows that the EWC and SI have the worst ACCs performances compared with other baselines. Such poor performance might caused by inappropriate regularization, which restricts the ability of learning the new task when preserving prior knowledge. The sub-optimal LA performances also indicate the limited learning ability on new tasks. In addition, from the Figure \ref{fig:learning-curve}, the performance of SI drops dramatically when the number of learning tasks increases. One reason is that the fixed model size limits the learning capacity.

\subsection{Effect of the replay-based methods} 
From Table \ref{table:perf}, we find that the NR strategy performs best compared with other baselines, because the data replay utilizes the historical training samples when learning new tasks. BWT shows favorable results, indicating that NR encounters no forgetting issues. Compared with the NR strategy, the GEM strategy has a much smaller buffer size. When the buffer size increases from 512 to 1024, the GEM strategy did not get the higher ACC performances but brings more heavy system workloads. 
  
\subsection{Effect of the proposed PCL-KWS}
Table \ref{table:perf} shows that the PCL-KWS achieves the best ACC and LA performances with fewer \textit{Extra Param} compared with other competitive methods. Specifically, compared with the best regularization-based method (i.e., SI), PCL-KWS improves the ACC by 37.2\% with only 17.46K \textit{Extra Param}. Meanwhile, compared with the best replay method (i.e., NR), the PCL-KWS reaches 8.7\% ACC improvement with no buffer and 7-fold training time reduction. We then analyse the ACC performances of the proposed PCL-KWS with increasing numbers of tasks $T$. As shown in Figure \ref{fig:learning-curve}, the proposed PCL-KWS always maintains a high ACC compared with the other methods, indicating good immunity to catastrophic forgetting.

\begin{figure}[t]
\centering
\begin{subfigure}{0.235\textwidth}
  \label{fig:task_param}
  \includegraphics[width=42mm]{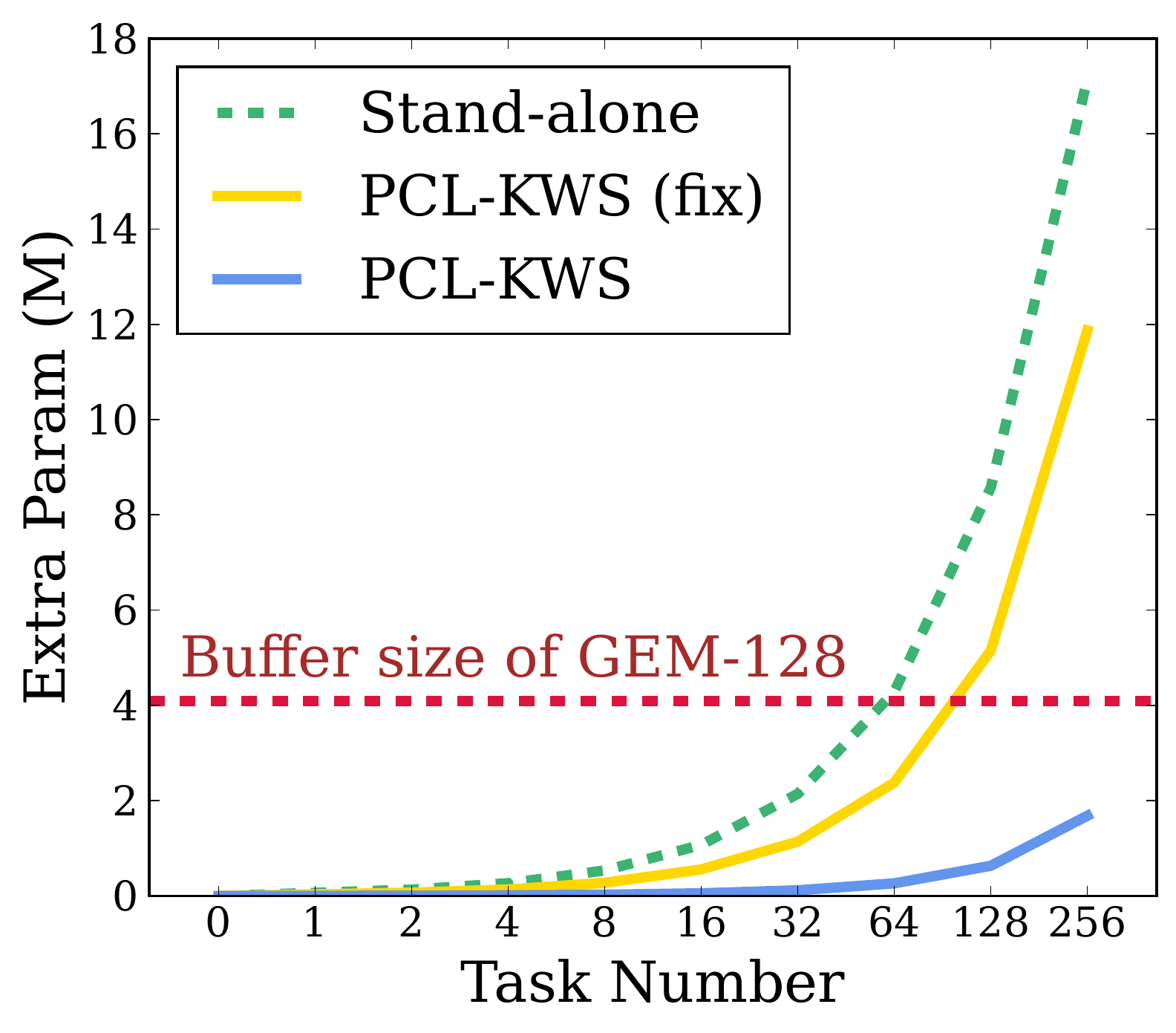} 
  \subcaption{}
\end{subfigure}
\begin{subfigure}{0.235\textwidth}  
    \label{fig:param_acc}
    \includegraphics[width=42mm]{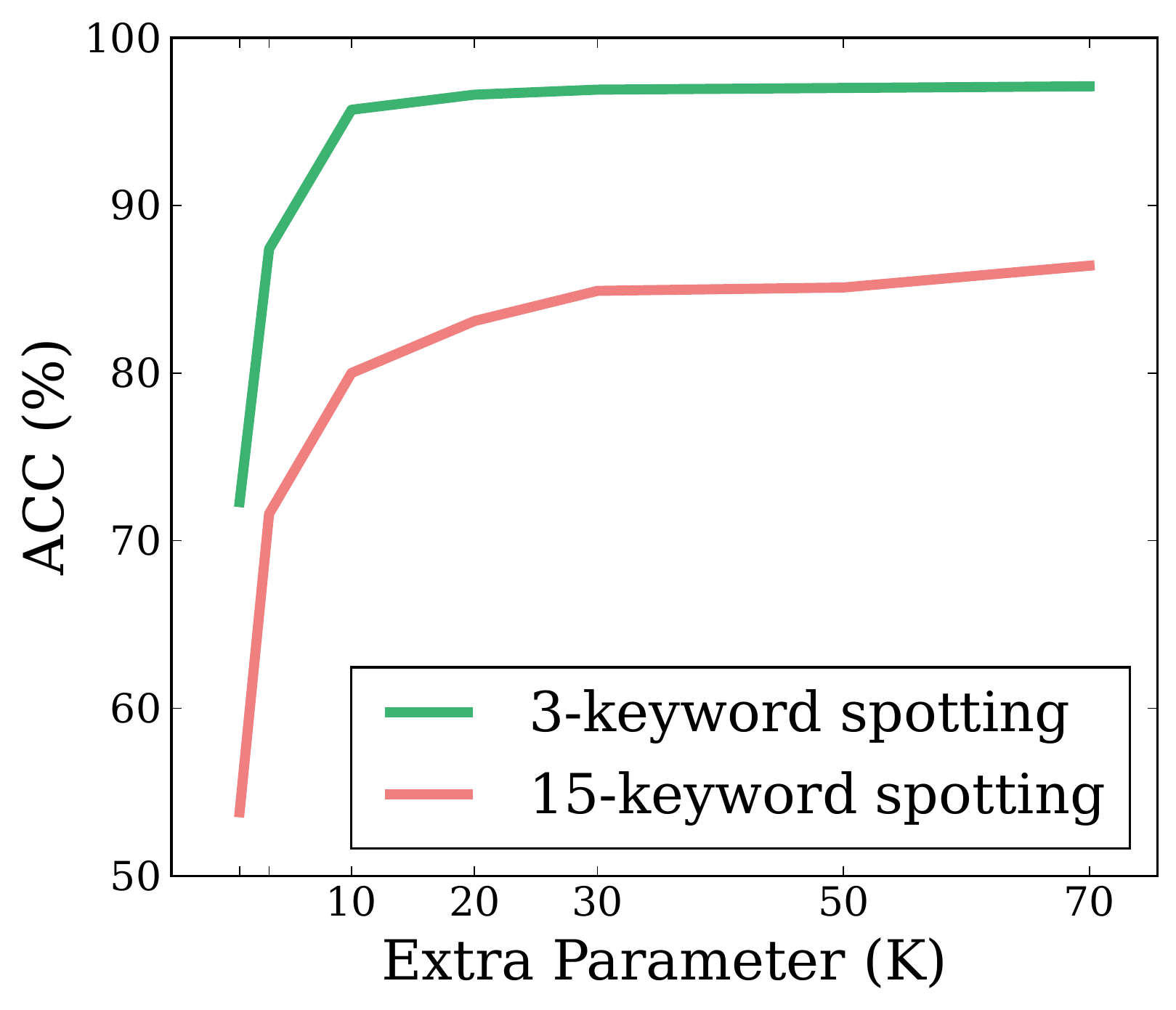}
    \subcaption{}
\end{subfigure}
\captionsetup{skip=5pt}
\caption{\textit{The extra parameter of PCL-KWS with the number of learned tasks (a), and the extra parameters with corresponding accuracy of the task with different keyword numbers (b).}}
\label{fig:efficiency}
\end{figure}

We further report the \textit{Extra Param} for the proposed PCL-KWS when there is increasing numbers of tasks $T$. As shown in Figure \ref{fig:efficiency}(a), even with $T=256$ tasks, the \textit{Extra Param} performances of the PCL-KWS is still less than 2M. This can be explained that the keyword-aware network scaling mechanism could significantly constrain the growth of the model parameters. In addition, we also illustrate the ACC performances of the proposed PCL-KWS with increasing \textit{Extra Param}. As shown in Figure \ref{fig:efficiency}(b), the ACC performance is first improved by the increasing \textit{Extra Param} but then he ACC curve saturates. Specifically, the proposed PCL-KWS requires at least 10K \textit{Extra Param} for 3-keyword spotting in each new task and at least 31K \textit{Extra Param} for 15-keyword spotting in each new task.

%%%%%%%%%%%%%%%%%%%%%%%%%%%%%%%%%%%%%%%%%%%%%%%%%%%%%%%%%%%%%%%%
%%%%%%%%%%%%%%%%%% Parameter and Accuracy Figure %%%%%%%%%%%%%%%
%%%%%%%%%%%%%%%%%%%%%%%%%%%%%%%%%%%%%%%%%%%%%%%%%%%%%%%%%%%%%%%%

\vspace{-12pt}
\section{Conclusion}
\label{sec:conclusion}

In this paper, we first explore four continual learning (CL) strategies (EWC, SI, GEM, NR) for KWS keyword incremental learning. Then, we proposed a progressive CL strategy (PCL-KWS) to alleviate the problem of catastrophic forgetting. Experimental results show that the proposed PCL-KWS archives the new state-of-the-art performance of 92.8\% average accuracy for all the tasks compared with other competitive baselines.

% Below is an example of how to insert images. Delete the ``\vspace'' line,
% uncomment the preceding line ``\centerline...'' and replace ``imageX.ps''
% with a suitable PostScript file name.
% -------------------------------------------------------------------------
% \begin{figure}[htb]

% \begin{minipage}[b]{1.0\linewidth}
%   \centering
%   \centerline{\includegraphics[width=8.5cm]{image1}}
% %  \vspace{2.0cm}
%   \centerline{(a) Result 1}\medskip
% \end{minipage}
% %
% \begin{minipage}[b]{.48\linewidth}
%   \centering
%   \centerline{\includegraphics[width=4.0cm]{image3}}
% %  \vspace{1.5cm}
%   \centerline{(b) Results 3}\medskip
% \end{minipage}
% \hfill
% \begin{minipage}[b]{0.48\linewidth}
%   \centering
%   \centerline{\includegraphics[width=4.0cm]{image4}}
% %  \vspace{1.5cm}
%   \centerline{(c) Result 4}\medskip
% \end{minipage}
%
% \caption{Example of placing a figure with experimental results.}
% \label{fig:res}
% %
% \end{figure}

%\vfill
%\pagebreak
\ninept

\bibliographystyle{IEEEbib}
\bibliography{icassp_final}

\end{document}